\pdfoutput=1

\documentclass[11pt]{article}

\usepackage{acl}

\usepackage{times}
\usepackage{latexsym}
\usepackage{graphicx}
\usepackage{amssymb}
\usepackage{amsmath}
\usepackage{booktabs}
\usepackage{multirow}
\usepackage{url}

\usepackage[T1]{fontenc}

\usepackage[utf8]{inputenc}

\usepackage{microtype}

\usepackage{inconsolata}

\makeatletter
\renewcommand{\@fnsymbol}[1]{%
  \ifcase#1\or \dagger\else\@ctrerr\fi}
\makeatother

\title{FastFiD: Improve Inference Efficiency of Open Domain Question Answering via Sentence Selection}

\author{Yufei~Huang$^{1,2}$\ \ \ \ Xu~Han$^{1,2}$\ \ \ \ Maosong Sun$^{1,2,3}$\thanks{\ \  Corresponding author.} \\
    $^1$Dept. of Comp. Sci. \& Tech., Institute for AI, Tsinghua University, Beijing, China \\
    $^2$Beijing National Research Center for Information Science and Technology \\
    $^3$Jiangsu Collaborative Innovation Center for Language Ability, Xuzhou, China \\
        \texttt{huang-yf20@mails.tsinghua.edu.cn} \ \ \texttt{\{hanxu2022,sms\}@tsinghua.edu.cn}
}

\begin{document}
\maketitle
\begin{abstract}
Open Domain Question Answering (ODQA) has been advancing rapidly in recent times, driven by significant developments in dense passage retrieval and pretrained language models. Current models typically incorporate the FiD framework, which is composed by a neural retriever alongside an encoder-decoder neural reader. In the answer generation process, the retriever will retrieve numerous passages (around 100 for instance), each of which is then individually encoded by the encoder. Subsequently, the decoder makes predictions based on these encoded passages. Nevertheless, this framework can be relatively time-consuming, particularly due to the extensive length of the gathered passages. To address this, we introduce FastFiD in this paper, a novel approach that executes sentence selection on the encoded passages. This aids in retaining valuable sentences while reducing the context length required for generating answers. Experiments on three commonly used datasets (Natural Questions, TriviaQA and ASQA) demonstrate that our method can enhance the inference speed by \textbf{2.3X-5.7X}, while simultaneously maintaining the model's performance. Moreover, an in-depth analysis of the model's attention reveals that the selected sentences indeed hold a substantial contribution towards the final answer. The codes are publicly available at \url{https://github.com/thunlp/FastFiD}.
\end{abstract}

\section{Introduction}

Open Domain Question Answering(ODQA) is a longstanding task in Natural Language Processing that involves generating an answer solely based on a given question. Recent advancements in this field have typically adopted the Retriever-Reader framework~\citep{chen-etal-2017-reading, karpukhin-etal-2020-dense, rag-neurips-2020, izacard-grave-2021-leveraging}, which breaks down the task into two distinct stages. Initially, a retriever retrieves a set of relevant passages from a high-quality collection of open domain documents, such as Wikipedia. Subsequently, a reader model generates an answer by considering the question and the retrieved passages. Thanks to advancements in neural models, the retriever has transitioned from traditional search methods like TF-IDF~\citep{chen-etal-2017-reading} to dense passage retrieval~\citep{karpukhin-etal-2020-dense}, resulting in improved retrieval performance. Furthermore, driven by the progress of Pretrained Language Models (PLMs)~\citep{devlin-etal-2019-bert, t5-jmlr, gpt3-neurips}, the reader has evolved from extracting answers from a single passage to generating answers from multiple passages~\citep{izacard-grave-2021-leveraging}. This approach enables the model to leverage information from various passages more effectively, thereby producing more accurate answers.

\begin{figure}
    \centering
    \includegraphics[width=0.48\textwidth]{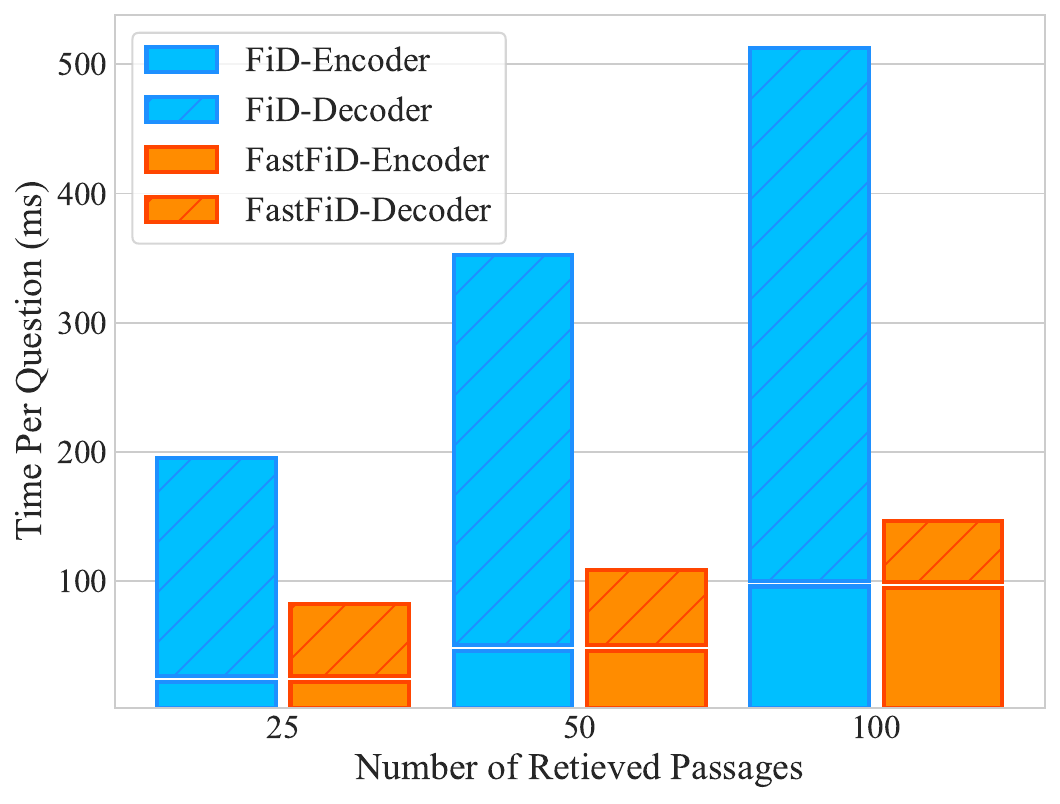}
    \caption{Inference Time for FiD (base) and FastFiD (base) with varying numbers of retrieved passages. As the number of retrieved passages increases, FiD encounters increasingly severe efficiency issues. Our FastFiD significantly accelerates the process by greatly reducing decoding time.}
    \label{fig:fid-inference-speed}
\end{figure}

A recently successful model is Fuse-in-Decoder (FiD)~\citep{izacard-grave-2021-leveraging}, which utilizes Dense Passage Retrieval and a generative reader based on T5~\citep{t5-jmlr}, an encoder-decoder model. FiD is capable of encoding each retrieved passage independently and subsequently concatenating these encoded passages to form an extensive context. The concatenated context is then used by the decoder to generate a response. Owing to its straightforward and extensible architecture, numerous subsequent works have introduced modifications based on this framework~\citep{sachan2021endtoend, yu-etal-2022-kg, wen-etal-2022-m3}. However, as the decoder must generate a response based on all retrieved passages, it can be time-consuming to enhance performance through the retrieval of additional passages. Moreover, in real-world scenarios, the latency in generating an answer is a significant factor. As larger language models continue to be developed and demonstrate superior performance, this issue may become more pronounced.

To address this issue, we introduce FastFiD, a novel approach that performs sentence selection post the encoder's output and maintains only the essential sentences as references for the decoder, thereby significantly reducing the inference time for each query.

To demonstrate the effectiveness of our approach, we first carry out experiments to ascertain that the multi-task training, which involves sentence selection and answer generation, does not conflict with one another during the model's learning process. This is achieved by seamlessly incorporating a selection loss on the encoder outputs with a language modelling loss on answer generation, enabling the model to simultaneously handle both sentence selection and answer generation tasks. An in-depth analysis of the decoder's cross-attention reveals that tokens from the chosen sentences yield a higher average attention score compared to those unchosen. This finding provides compelling evidence that the selected sentences significantly contribute more to the model's predictions. Guided by this insight, we execute a secondary training phase, obliging the model to solely anchor to the selected encoder outputs when making the final prediction.

The experimental results obtained from two widely used ODQA datasets, namely Natural Questions (NQ)~\citep{kwiatkowski-etal-2019-natural} and TriviaQA~\citep{joshi-etal-2017-triviaqa}, along with a long-form QA dataset called ASQA~\citep{hofstatter2023fid}, demonstrate that FastFiD can achieve performance metrics comparable to the original FiD. Notably, it can reduce the context length by up to \textbf{38X} and accelerate the inference time by \textbf{2.3X-5.7X} on different datasets. To validate the effectiveness of sentence selection, we also compare its performance with passage reranking after the encoder outputs. The results show that sentence selection yields better performance while maintaining a similar context length. This comparison indicates that sentence selection is a more effective strategy for compressing information across multiple passages.

In summary, our contributions can be encapsulated within the following three key points:
\begin{itemize}
    \item We implement a multi-task training approach, demonstrating that a singular reader model can concurrently perform sentence selection and answer generation.
    \item We introduce a novel technique to enhance the inference efficiency of FiD while preserving its question-answering capabilities.
    \item We carry out plenty of experiments to validate and analyze the effectiveness of our method.
\end{itemize}

\section{Related Work}

\begin{figure*}[ht]
    \centering
    \includegraphics[width=\linewidth]{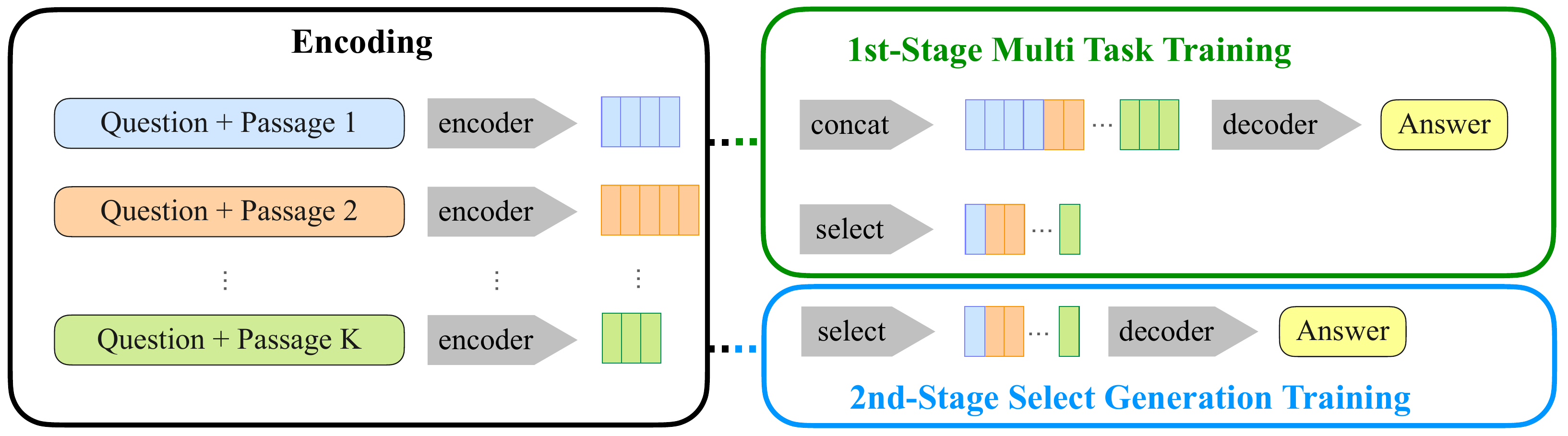}
    \caption{An overview of our FastFiD training pipeline. The pipeline undergoes two stages of training to empower the model with the capacity to generate answers based on the selected sentences, thereby minimizing inference time.}
    \label{fig:model}
\end{figure*}

\paragraph{Open Domain Question Answering} serves a crucial role in natural language processing, with its primary function being to respond to factoid questions.  Followed by \citet{chen-etal-2017-reading}, current ODQA systems usually use a large collection of documents like Wikipedia as the knowledge source to answer questions. Since the document collection usually contains millions of documents, the system always adds a retriever to retrieve some most relevant passages for the reader to make predictions. To get better retriever performance, \citet{karpukhin-etal-2020-dense} proposed a shift from sparse retrieval systems like TF-IDF to dense retrieval to enhance the efficiency of the retriever. Subsequent research~\citep{rag-neurips-2020, sachan2021endtoend, jiang-etal-2022-retrieval, lee-etal-2022-need} has investigated the use of end-to-end training methodologies to further boost the performance of the retriever, bypassing the need for pair-wise question-document data. \citet{izacard2021distilling} demonstrated an improvement in performance through the distillation of knowledge from the reader to the retriever. The idea of pretraining both the retriever and the reader on a vast, unlabeled corpus has been explored by \citet{pmlr-v119-guu20a} and \citet{sachan-etal-2021-end}. A different research trajectory has aimed to augment the reader's capacity to better utilize retrieved passages. With the advancement of PLMs, the reader has evolved from RNN-based models~\citep{chen-etal-2017-reading} to BERT-based extractive readers ~\citep{karpukhin-etal-2020-dense} and T5 or BART-based generative readers~\citep{rag-neurips-2020, izacard-grave-2021-leveraging}. Recent studies~\citep{cheng-etal-2021-unitedqa, fajcik-etal-2021-r2-d2, wen-etal-2022-m3} have pivoted towards a hybrid approach, exploring the integration of both generative and extractive readers to further enhance system performance.

\paragraph{Efficient ODQA} The majority of contemporary Open-Domain Question Answering (ODQA) systems face efficiency challenges, primarily due to the large-scale document processing and the use of sizable pre-trained language models. These efficiency challenges arise in two stages.

The first stage is retrieval efficiency. Given the potentially massive number of passages, dense retrieval can be extremely slow. Instead of relying solely on brute force search methods, alternative algorithms such as Approximate Nearest Neighbor (ANN)~\citep{johnson-ann-2021} and Hierarchical Navigable Small World (HNSW)~\citep{malkov-hsnw-2020} can be employed to expedite the retrieval process.

The second efficiency challenge lies in the reading process, which involves handling multiple passages for each query. To address this, \citet{hofstatter2023fid} propose FiD-Light, which limits the decoder's attention to the first k tokens of each passage to reduce the context length. FiDO~\citep{de-jong-etal-2023-fido} explores reducing the number of cross attention layers in FiD's decoder to increase efficiency, but this comes at the cost of re-pretraining the base model. Other complementary strategies explore to identify and stop processing less relevant passages early on by utilizing adaptive computation~\citep{wu-etal-2020-dont, wu-etal-2021-training} or knowledge graph with GNN network~\citep{yu-etal-2022-kg}. Additionally, some research has focused on directly retrieving answers to questions without the need for passage processing~\citep{seo-etal-2019-real, lee-etal-2021-learning-dense, lewis-paq}, or using language models to generate answers directly by finetuning and few-shot prompting~\citep{roberts-etal-2020-much, gpt3-neurips}. 

\paragraph{Answer Sentence Selection} Answer Sentence Selection (AS2) is a long-standing task that has been extensively explored. Dense Neural Networks (DNNs) have been widely employed in this task~\citep{10.1145/2766462.2767738, shen-etal-2017-inter}. \citet{Garg_Vu_Moschitti_2020} further advanced the field by utilizing transformer-based pre-trained language models (PLMs) to achieve better results. Recent studies have investigated methods such as generating answer sentences~\citep{hsu-etal-2021-answer} and implementing complex ranking pipelines~\citep{10.1145/3397271.3401266}. Unlike these approaches, our work aims to predict the exact answer span from retrieved passages, using answer sentence selection only for enhancing inference speed.

\section{Methods}

In this section, we propose FastFiD, which is based on FiD~\citep{izacard-grave-2021-leveraging} to reduce its inference time and make it more efficient. FastFiD contains a two-stage training procedure. Initially, in the first stage, we introduce a multi-task training objective that allows for simultaneous training of sentence selection and answer generation~(Section~\ref{subsec:mutl-task-training}). Then, in the second stage, we use the model trained in the first stage as the base model and perform continuous training on generating answers with reference to the selected tokens. (Section~\ref{subsec:select-generation-training}). Finally, in the inference stage, the encoder transcodes each passage into context embeddings and curates a selection of valuable sentences, which are then employed in the decoder generation process to expedite inference time (Section~\ref{subsec:select-generation-inference}). The overall framework is shown in Figure~\ref{fig:model}.

\subsection{Multi-Task Training}
\label{subsec:mutl-task-training}
In this section, we present our multi-task training approach. Following FiD, we utilize T5, an encoder-decoder based PLM, as our base model. Given a question-answer pair $(q, a)$, we initially retrieve $K$ relevant passages ${p^1, p^2, ..., p^K}$, with their respective titles ${t^1, t^2, ..., t^K}$ from an extensive knowledge base, predicated on the question $q$. Subsequently, the question $q$ and each corresponding passage $p^k$ are combined to generate a comprehensive input in the following structure:
\begin{equation}
I^k = \mathrm{\textbf{Question:}}\ q\ \mathrm{\textbf{Title:}}\ t^k\ \mathrm{\textbf{Context:}}\ p^k
\end{equation}

After this, the model's encoder transcodes each input $I^k$ into context embeddings ${h_1^k, h_2^k, ..., h_N^k} \in \mathbb{R}^d$, where $N$ represents the max sequence length of the input text. Our multi-task training objective, which encompasses sentence selection and answer generation, is built upon these encoded context embeddings.

\subsubsection{Sentence Selection}

In the context of a given retrieved passage $p^k$, there exist $M_k$ key sentences, represented as $\mathcal{S}^k = {s^k_{1}, s^k_2, ..., s^k_{M_k}}$, that are crucial for answering the question. As established in prior extractive reader works~\citep{chen-etal-2017-reading, kwiatkowski-etal-2019-natural, min-etal-2019-discrete, cheng-etal-2021-unitedqa}, we implement a classification head to anticipate the begin and end positions of each key sentence. Taking into account the conclusions of \citet{cheng-etal-2020-probabilistic} and \citet{cheng-etal-2021-unitedqa}, we employ a multi-objective approach to enhance sentence selection performance.

In formal terms, the probability of a span $(i^k, j^k)$ being a selected sentence can be broken down into the product of the probabilities of the $i^k$-th token being the start token and the $j^k$-th token being the end token. We integrate some learned parameters, namely $w_b, w_e, b_b, b_e$, to calculate the start and end score:

\begin{equation}
    \begin{aligned}
        S_b(i^k) &= w_b^T h_i^k + b_b; \\
        S_e(j^k) &= w_e^T h_j^k + b_e \\
    \end{aligned}
\end{equation}

By calculating the probability based on different normalizing factors, we can derive the local passage-level probability and the global multi-passage-level probability. With local probability, the probability of each token in different retrieved passages will not affect one another. By normalizing the start and end probabilities by the total scores of all tokens in input $I^k$, we derive the probability as follows:
\begin{equation}
    \begin{aligned}
        P^L_b(i^k) &= \frac{\exp(S_b(i^k))}{\sum_{n}\exp(S_b(n^k))}; \\
        P^L_e(j^k) &= \frac{\exp(S_e(j^k))}{\sum_{n}\exp(S_e(n^k))}
    \end{aligned}
\end{equation}

In the case of global probability, we calculate the probability taking into account all the tokens in the top-K passages from the retriever. Therefore, the probability of each token being the start or end of the selected sentence will be jointly optimized across different passages:

\begin{equation}
    \begin{aligned}
        P^G_b(i^k) &= \frac{\exp(S_b(i^k))}{\sum_{k}\sum_{n}\exp(S_b(n^k))}; \\
        P^G_e(j^k) &= \frac{\exp(S_e(j^k))}{\sum_{k}\sum_{n}\exp(S_e(n^k))}
    \end{aligned}
\end{equation}

We then obtain the local and global probabilities of a span being the supported sentence as follows:

\begin{equation}
    P^{\{L, G\}}_s(i^k, j^k) = P^{\{L, G\}}_b(i^k) \times P^{\{L, G\}}_e(j^k)
\end{equation}

Following the methodology of \citet{cheng-etal-2021-unitedqa}, we utilize a multi-objective formulation to merge the HardEM~\citep{min-etal-2019-discrete} and MML~\citep{karpukhin-etal-2020-dense} objectives for more efficient training. In the multi-objective formulation, we calculate the HardEM loss on global probability and the MML loss on local probability. The final sentence selection loss is calculated as follows:
\begin{equation}
    \begin{aligned}
    \mathcal{L}_{S} =& -\log \max_{(i, j) \in \mathcal{S}} P^G_s(i, j) - \\
                     & \frac{1}{K} \sum_{k}^K \log \sum_{(i^k, j^k) \in \mathcal{S}^k }P^L_s(i^k, j^k)
    \end{aligned}
\end{equation}
where $\mathcal{S} = \mathcal{S}^1 \cup \mathcal{S}^2 \cup ...\cup \mathcal{S}^K$ is the set of all crucial sentences in the top-K retrieved passages. Since ODQA datasets usually only contain question-answer pairs without annotated valuable sentences, we consider the sentences that include the short span answer in each retrieved passage as the crucial sentences.

\subsubsection{Answer Generation}

As the pipeline in FiD, we employ the decoder to fuse the information of retrieved passages and make a prediction. More specifically, we first concatenate the context embeddings of all inputs:
\begin{equation}
    H = (H^1; H^2; ...; H^K)
\end{equation}
where $H^k$ represents the context embeddings for input $I^k$, therefore $H$ have an overall length of $N \times K$. Subsequently, the decoder conducts cross-attention over the concatenated context embeddings to make generation.

For the training objective, it optimizes the language modelling loss of generating the golden answer $a$, a sequence of tokens represented as $\{a_1, a_2, ..., a_{N_a}\}$:
\begin{equation}
    \mathcal{L}_G = -\log \sum_{i}^{N_a} P_{\theta_d}(a_i|H, a_{1:i-1})
\end{equation}
where $\theta_d$ is parameters of the decoder.

Finally, in the first-stage multi-task training, we integrate the sentence selection objective and answer generation objective in the following manner to simultaneously equip the model with these two capabilities. The variable $\lambda$ is a hyper-parameter that balances these two objectives:
\begin{equation}
    \mathcal{L}^1 = \mathcal{L}_G + \lambda \mathcal{L}_S
\end{equation}

\subsection{Select Generation Training}
\label{subsec:select-generation-training}
After completing the initial stage of training as outlined in Section~\ref{subsec:mutl-task-training}, our preliminary experiments reveal that while the model possesses the capacity to select valuable sentences and make predictions at the same time, directly requiring the decoder to form predictions solely based on these selected sentences significantly hampers the performance of the model. We hypothesise that this is because of the gap in context length for decoder between training and inference. Therefore, we introduce a second stage of continuous training aimed at minimizing this discrepancy linked with context length.

More specifically, we initially obtain the context embeddings of the selected sentences, and this is done by the global multi-passage-level selection probability.
\begin{equation}
\begin{aligned}
    &H_s = \bigcup h_{i^k:j^k}; \\
    &(i^k, j^k) \in \mathrm{TopK}(P^G_s(i, j))
\end{aligned}
\end{equation}

The resultant loss for answer generation can then be expressed as follows:
\begin{equation}
    \mathcal{L}_G^s = -\log \sum_{i}^{N_a} P_{\theta_d}(a_i|H_s, a_{1:i-1})
\end{equation}

Throughout the second stage of training, we maintain the use of a multi-task training objective to keep both the sentence selection ability and answer generation ability, thereby facilitating better performance.
\begin{equation}
    \mathcal{L}^2 = \mathcal{L}_G^s + \lambda \mathcal{L}_S
\end{equation}

\subsection{Select Generation Inference}
\label{subsec:select-generation-inference}

Following the two-stage training process, we acquire a model that is capable of dynamically selecting valuable sentences for the decoder to make generation. The inference process closely mirrors the second stage of training described in Section~\ref{subsec:select-generation-training}. Initially, valuable context embeddings are selected based on global selection probability. Subsequently, a greedy decoding strategy is employed to generate the answer based on the selected context embeddings denoted as $H_s$.

\section{Experiments}

\subsection{Experimental Setup}

Same as FiD~\citep{izacard-grave-2021-leveraging}, we utilize T5~\citep{t5-jmlr} as our base model. For passage retrieval, we utilize the retriever demonstrated by \citet{izacard2021distilling} which has superior retrieval performance. Following previous work~\citep{lee-etal-2019-latent, karpukhin-etal-2020-dense}, we use the preprocessed English Wikipedia Snapshot on 12-30-2018 as our knowledge source. And we use average time per question (TPQ) to measure model's inference efficiency. We conduct experiments on two commonly used ODQA datasets and one long-form QA dataset. Their statistics are shown in Table~\ref{tab:statistics}. We use the original train/dev/test split to conduct our experiments.

\begin{table}[t]
    \centering
    \small
    \begin{tabular}{c|cccc}
        \toprule
        \ & \#Train & \#Dev & \#Test & \#Sent. \\
         \midrule
        NQ & 79,168 & 8,757 & 3,610 & 14.84 \\
        TriviaQA & 76,423 & 8,837 & 11,313 & 30.58 \\
        ASQA & 4,353 & 968 & 1,015 &  22.32 \\
        \bottomrule
    \end{tabular}
    \caption{Statistics of two ODQA datasets. \#Train/\#Dev/\#Test imply the number of train/dev/test samples. \#Sent. means the average number of valuable sentences recognized in top-100 retrived passages.}
    \label{tab:statistics}
\end{table}

\paragraph{Natural Questions}\citep{kwiatkowski-etal-2019-natural} is a large ODQA dataset where all questions are mined from Google Search real queries. The annotated answers are all created by human annotators based on Wikipedia documents. \citet{lee-etal-2019-latent} further filter out questions with short answers to construct the open domain version of NQ, which we used in our experiment. We evaluate the performance of our model on NQ using the Exact Match (EM) metric.

\paragraph{TriviaQA}\citep{joshi-etal-2017-triviaqa} is collected from 14 trivia and quiz-league websites with human-annotated answers and a set of answer aliases gathered from Wikipedia. We use the unfiltered question-answer pairs and discard the distantly supervised documents as our open domain version. Similar to NQ, we assess our model's performance on TriviaQA using the Exact Match (EM) metric.

\paragraph{ASQA}\citep{stelmakh2023asqa} is a long-form question answering dataset that builds upon the AmbigQA~\citep{min-etal-2020-ambigqa} dataset. It consists of ambiguous questions with multiple short span answers and long-form answers from human annotators that coverage all possible short span answers. In line with \citet{stelmakh2023asqa}, we evaluate the performance of our model on this dataset using the STR-EM (String Exact Match) metric. STR-EM measures the proportion of disambiguated short answers that are correctly identified within the long answer. Since the test set of ASQA is not publicly available, our evaluation is conducted solely on the development set of ASQA.

\paragraph{Baselines} We mainly compare our method with vanilla FiD, aiming at enhancing its inference efficiency. Additionally, we contrast our approach with the model resulting from our first training stage, referred to as HybridFiD, a model that is capable of simultaneously performing answer generation and sentence selection. Besides, we also compare with FiD-Light~\citep{hofstatter2023fid}, which propose to select the first-k tokens from each passage as the context for decoder and improve efficiency.

\paragraph{Implementation} Our method is implemented using PyTorch~\citep{adam-etal-2019-pytorch} and Huggingface Transformers~\citep{wolf-etal-2020-transformers}, with training efficiency enhanced by DeepSpeed ZeRO-2~\citep{rajbhandari2020zero}. Due to GPU limitations, we conduct experiments using T5-Base, which has 345M parameters. We employ the AdamW~\citep{DBLP:conf/iclr/LoshchilovH19} optimizer for stable training. More implementation details are shown in Appendix~\ref{sec:implemenration}.

\subsection{Main Results}
\label{subsec:main-results}

\begin{table*}[!t]
\centering
\begin{tabular}{lccccccccc}
\toprule
\multirow{2}{*}{\textbf{Model}} & \multicolumn{3}{c}{\textbf{NQ}}   & \multicolumn{3}{c}{\textbf{TriviaQA}} & \multicolumn{3}{c}{\textbf{ASQA}}  \\
 & EM    & TPQ & Speed & EM   & TPQ & Speed & STR-EM & TPQ & Speed \\ \midrule[1pt]
FiD       & 50.06          & 514  & 1.0X      & \textbf{69.79}  & 550 & 1.0X     & 33.35   & 3,323  & 1.0X \\
FiD-Light & 40.91          & 201  & 2.6X    & 63.15  & \textbf{218}   & \textbf{2.5X}  & 27.34   & 867  & 3.8X \\ \hline
HybridFiD & 50.14          & 513  & 1.0X      & 69.77  & 540 & 1.0X   & 35.13           & 3,330  & 1.0X  \\
FastFiD   & \textbf{50.17} & \textbf{148} & \textbf{3.5X} & 69.34 & 241 & 2.3X & \textbf{37.22} & \textbf{586} & \textbf{5.7X}    \\ \bottomrule
\end{tabular}
\caption{Performance of vanilla FiD, FiD-Light, HybridFiD, FastFiD with 100 retrieved passages on test set (development set for ASQA). We select 200 sentences for NQ and ASQA, 400 sentences for TriviaQA. For FiD-Light, we utilize a value of 64 for $k$, which as demonstrated by \citet{hofstatter2023fid}, yields the best performance. TPQ is measured by milliseconds.}
\label{tab:main_results}
\end{table*}

\begin{table*}[t]
    \centering
    \begin{tabular}{l|cccccc}
    \toprule
    \textbf{Model} & \textbf{\# Doc} & \textbf{NQ-Dev} & \textbf{NQ-Test} & \textbf{Context Length} & \textbf{TPQ} & \textbf{Speed} \\ \midrule[1pt]
    FiD            & 25              & 47.33           & 47.23            & 9,600                    & 197                   & 1.0X   \\
    HybridFiD      & 25              & \textbf{47.71}  & \textbf{48.42}   & 9,600                    & 194                   & 1.0X   \\
    FastFiD        & 25              & 47.52           & 48.06            & 920                     & 84                   & \textbf{2.4X}   \\ \hline
    FiD            & 50              & 47.79           & 47.89            & 19,200                   & 354                   & 1.0X   \\
    HybirdFiD      & 50              & \textbf{48.12}   & \textbf{49.09}   & 19,200                   & 354                   & 1.0X   \\
    FastFiD        & 50              & 47.96           & 48.89   & 1,035                    & 110                   & \textbf{3.2X}   \\ \hline
    FiD            & 100             & \textbf{49.10}  & 50.06   & 38,400                   & 514                   & 1.0X   \\
    HybirdFiD      & 100             & 48.65           & 50.14           & 38,400                   & 513                   & 1.0X   \\
    FastFiD        & 100             & 48.98           & \textbf{50.17}            & 1,008                     & 148                   & \textbf{3.5X}   \\ \bottomrule
    \end{tabular}
    \caption{Detailed performance of vanilla FiD, HybridFiD and FastFiD on NQ with different number of passages.}
    \label{tab:nq_results}
\end{table*}

\begin{figure}[t]
    \centering
    \includegraphics[width=\linewidth]{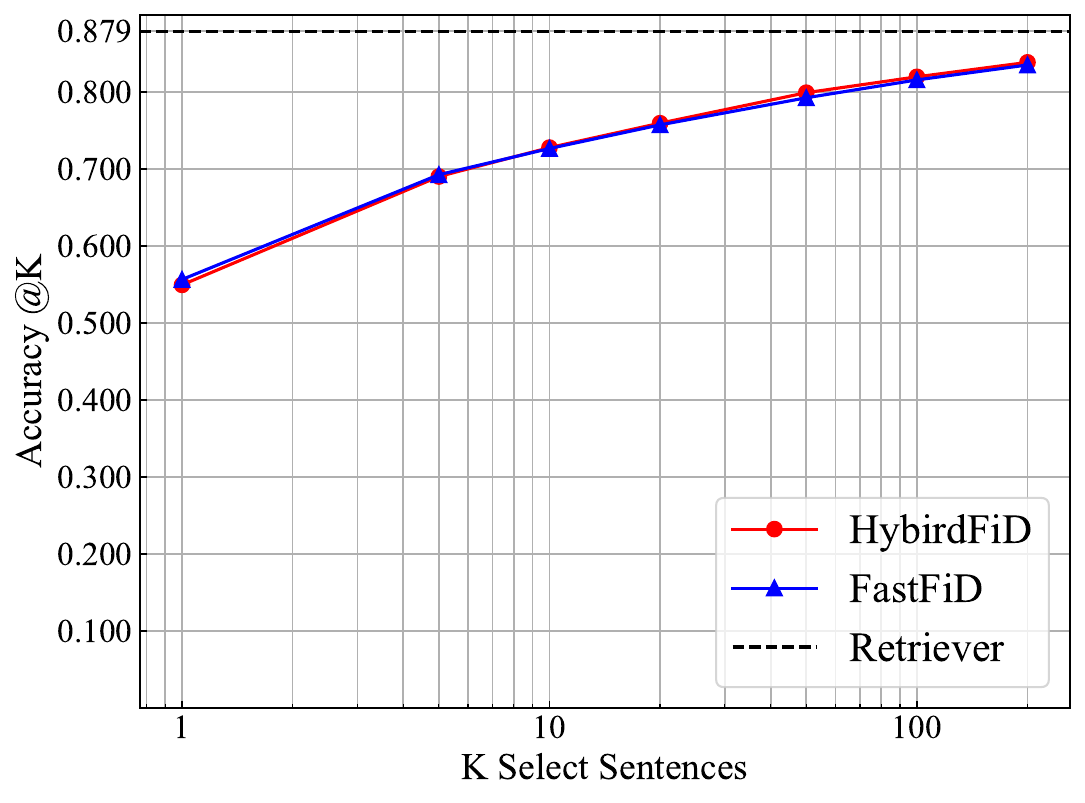}
    \caption{Sentence selection performance on NQ-Dev for HybirdFiD and FastFiD with 100 retrieved passages. Retriever means the accuracy of our retriever when retrieving 100 passages, which can be seen as an upper bound.}
    \label{fig:extraction-accuracy}
\end{figure}

\paragraph{Answer Generation} The performance and inference speed of our FastFiD and other baselines are presented in Table~\ref{tab:main_results}. Unlike FiD-Light, which sacrifices QA performance to accelerate the inference process, FastFiD achieves substantial acceleration while maintaining similar or even superior QA performance compared to vanilla FiD. Additionally, FastFiD demonstrates significantly greater inference speedup than FiD-Light on NQ and ASQA, and comparable acceleration on TriviaQA. This can be attributed to our context-aware compression methods, which extract more essential information with fewer tokens compared to the static method employed in FiD-Light. Among the three datasets, FastFiD achieves the highest acceleration on ASQA due to the longer answer format. This showcases the effectiveness of FastFiD in long-form QA, which is a widely utilized task by modern LLM system like New Bing\footnote{\url{https://www.bing.com/}} and ChatGPT\footnote{\url{https://chat.openai.com/}}.

We also conducted experiments with varying numbers of retrieved passages on NQ, and the results are presented in Table~\ref{tab:nq_results}. As observed, regardless of the number of retrieved passages, our FastFiD consistently matches or even surpasses FiD and HybridFiD in terms of EM, while significantly reducing the context length and inference time. Moreover, as the number of retrieved passages increases, the speedup rate also expands. This evidence underscores the potential of our method for effective implementation with a larger number of passages or lengthy documents.

\paragraph{Sentence Selection} Similar to the metrics employed in the retriever, we measure the performance of sentence selection utilizing the accuracy@k, which assesses whether the correct answer appears within the top-k sentences. As depicted in Figure~\ref{fig:extraction-accuracy}, there is a positive correlation between the increase in selected sentence numbers and accuracy, eventually surpassing 95\% of the retriever's accuracy for both HybridFiD and FastFiD. This demonstrates their substantial capability to select valuable sentences. A comparative evaluation of FastFiD and HybridFiD indicates that the second-stage training has a minimal impact on the sentence selection performance. Its main contribution is to adapt the model to the reduced context length, as we anticipated.

\paragraph{Discussion} The performance of HybridFiD, as presented in Table~\ref{tab:main_results} and Figure~\ref{fig:extraction-accuracy}, highlights that answer generation and sentence selection are not mutually exclusive, and a multi-task training objective enables both capabilities. To further explore the relationship between sentence selection and answer generation, we examined the average cross-attention scores for tokens within the top 200 sentences and the non-selected segments. This analysis was conducted using HybridFiD with 100 retrieved passages on NQ. Following the approach of \citet{izacard2021distilling}, we calculated the cross-attention score of each token in the inputs by averaging across all decoder layers, attention heads per layer, and all generated tokens.

Table~\ref{tab:cross_attention} shows that the selected sentences have significantly higher average cross-attention scores compared to the non-selected segments, indicating that they contribute more significantly to the final answer generation. Conversely, this suggests that the non-selected segments largely contain irrelevant information, contributing less to answer generation despite being present in the context, and can therefore be disregarded during the decoding process. This insight also served as a motivation for our second-stage training, as described in Section~\ref{subsec:select-generation-training}. Furthermore, for a more comprehensive understanding of the effectiveness of our FastFiD approach, we provide a detailed case study in Appendix~\ref{sec:case_study}.

\begin{table}[ht]
    \centering
    \begin{tabular}{l|cc}
    \toprule
        \ & NQ-Dev & NQ-Test \\ \midrule
        Selected     & \textbf{5.28E-4} \ & \textbf{5.32E-4} \\
        Non-Selected & 3.46E-5 \ & 3.43E-5 \\ \bottomrule
    \end{tabular}
    \caption{Average cross-attention score for tokens in top-200 selected sentences and non-selected sentences for HybirdFiD with 100 retrieved passages.}
    \label{tab:cross_attention}
\end{table}

\section{Further Analysis}

In this section, we present additional experiments to demonstrate the effectiveness of our method. First, we compare our sentence selection method with the passage reranking method in Section~\ref{subsec:rerank}. Second, we evaluate the performance of our method with varying numbers of selected sentences in Section~\ref{subsec:number_sentences}. Third, we conduct an ablation study to verify the importance of our two-stage training approach in Section~\ref{subsec:two_stage}. Finally, we assess the effectiveness of our method on decoder-only models in Section~\ref{subsec:llm_results}.

\subsection{Sentence Selection vs Passage Rerank}
\label{subsec:rerank}

Similar to conducting sentence selection after the encoder, another method is to conduct passage rerank after encoder's outputs and thus reducing context length and inference time. In alignment with our two-stage training pipeline, we substitute the sentence selection loss with a passage reranking loss as utilized by \citet{nogueira2020passage}, leading to a model we name RerankFiD. We evaluate the performance of FastFiD and RerankFiD under comparable context lengths, with the findings presented in Table~\ref{tab:rerank}. Consistently, our FastFiD method outperforms RerankFiD across a range of retrieved passage quantities. We hypothesize that this is due to the higher density of related information in the selected sentences compared to the reranked passages, as a passage often includes numerous irrelevant sentences even if it contains the correct answer.

\begin{table}[t]
    \centering
    \small
    \begin{tabular}{@{}l|cccc@{}}
    \toprule
\multicolumn{1}{c|}{\textbf{Model}} & \textbf{\# Doc} & \textbf{NQ-Dev} & \textbf{NQ-Test} & \textbf{\begin{tabular}[c]{@{}c@{}}Context\\ Length\end{tabular}} \\ \midrule
FastFiD        & 25              & \textbf{47.52} & \textbf{48.06} & 920                     \\
RerankFiD      & 25              & 46.42 & 47.20 & 1,152  \\ \hline
FastFiD        & 50              & \textbf{47.96} & \textbf{48.89} & 1,035                   \\
RerankFiD      & 50              & 46.64 & 47.23 & 1,152                    \\ \hline
FastFiD        & 100             & \textbf{48.98} & \textbf{50.17} & 1,008                     \\
RerankFiD      & 100             & 46.45 & 48.09 & 1,152  \\ \bottomrule
    \end{tabular}
    \caption{Comparison between FastFiD and RerankFiD among different number of retrieved passages. FastFiD consistently outperforms RerankFiD within similar context length.}
    \label{tab:rerank}
\end{table}

\subsection{Number of Selected Sentences}
\label{subsec:number_sentences}

To evaluate the impact of varying the number of selected sentences, we conducted experiments on NQ with 100 retrieved passages. The results in Table~\ref{tab:number_of_select} show that increasing the number of selected sentences leads to a nearly linear increase in the context length for the decoder. In terms of answer generation effectiveness, FastFiD performs well even with only 50 selected sentences and improves gradually with more sentences selected. It is worth noting that performance reaches a plateau after a certain number of sentences, such as 200. Beyond this point, selecting additional sentences does not yield further improvement but only increases context length and inference time.

\begin{table}[t]
\small
\begin{tabular}{@{}l|cccc@{}}
\toprule
\multicolumn{1}{c|}{\textbf{Model}} & \textbf{\begin{tabular}[c]{@{}c@{}}\# Select\\ Sentence\end{tabular}} & \textbf{NQ-Dev} & \textbf{NQ-Test} & \textbf{\begin{tabular}[c]{@{}c@{}}Context\\ Length\end{tabular}} \\ \midrule
FiD      & -    & \textbf{49.10}           & 50.06            & 38,400  \\ \hline
FastFiD  & 50  & 48.25 & 49.11 & 378   \\
FastFiD   & 100  & 48.29 & 49.28 & 639    \\
FastFiD  & 200 &  48.98 & \textbf{50.17} & 1,008  \\
FastFiD  & 400 & 49.05 & 49.83 & 1,661  \\ \bottomrule
\end{tabular}
\caption{Experiments on the number of selected sentences.}
\label{tab:number_of_select}
\end{table}

\subsection{Two-Stage Training}
\label{subsec:two_stage}

To corroborate the efficacy of our two-stage training approach, we undertake experiments wherein each training stage is separately removed, with the outcomes displayed in Table~\ref{tab:two_stage}. It is evident that the removal of either training stage results in a decrement in the final performance. Moreover, the second stage of training appears to be more consequential than the first stage, as demonstrated by the nearly 10-point drop in performance when the second stage is removed, compared to a decrease of less than 1-point when only the second stage is implemented.

\begin{table}[ht]
    \small
    \begin{tabular}{@{}l|cccc@{}}
    \toprule
    \multicolumn{1}{c|}{\textbf{Model}} & \textbf{\# Doc} & \textbf{\begin{tabular}[c]{@{}c@{}}\# Select\\ Sentence\end{tabular}} & \textbf{NQ-Dev} & \textbf{NQ-Test} \\ \midrule
    FastFiD                             & 50              & 200  & \textbf{47.96} & \textbf{48.89}           \\
    - 2nd-stage                         & 50              & 200  & 36.61 & 37.67            \\
    - 1st-stage                         & 50              & 200  & 47.62 & 48.03            \\ \hline
    FastFiD                             & 100             & 200  & \textbf{48.98} & \textbf{50.17}           \\
    - 2nd-stage                         & 100             & 200  & 38.62 & 39.25            \\
    - 1st-stage                         & 100             & 200  & 48.25 & 49.17            \\ \bottomrule
    \end{tabular}
    \caption{Ablation study on two-stage training method.}
    \label{tab:two_stage}
\end{table}

\subsection{Application on Decoder-Only LLM}
\label{subsec:llm_results}

With the success of ChatGPT and GPT-4~\citep{openai2024gpt4}, most large language models~\citep{DBLP:journals/corr/abs-2302-13971, DBLP:journals/corr/abs-2307-09288} are currently built on a decoder-only architecture and demonstrate superior performance in ODQA. Consequently, we conducted additional experiments to evaluate the effectiveness of our method on decoder-only models. To adapt these models, we made a single minor modification: the sentence selection head on top of the decoder now extracts the key-value caches of selected sentences instead of the final hidden states. These selected key-value caches are then utilized to accelerate the inference process.

Based on this, we conducted experiments on Llama2-7B~\citep{DBLP:journals/corr/abs-2307-09288} using the NQ dataset with 20 retrieved passages to verify our method, as shown in Table~\ref{tab:llama2_results}. The results demonstrate that our method can speed up Llama2-7B by \textbf{1.9} times, with only a minor decrease in performance. This acceleration is achieved by shortening the context length without losing important information, indicating potential for even greater speedup in future LLMs with longer sequences and more retrieved passages. Consequently, our method is well-suited to various architectures, providing a scalable way to enhance inference speed while maintaining performance.

\begin{table}[]
    \small
    \centering
    \begin{tabular}{l|ccc}
        \toprule
        Model     & EM & TPQ(ms) & Speed \\
        \midrule[1pt]
        Llama2-7B & 50.58 & 1,855 & 1.0X \\
        HybridLlama2-7B & \textbf{51.86} & 1,867 & 1.0X \\
        FastLlama2-7B & 48.95 & \textbf{966} & \textbf{1.9X} \\
        \bottomrule
    \end{tabular}
    \caption{Performance of Llama2, HybridLlama2 and FastLlama2 with 20 retrieved passages on test set of NQ.}
    \label{tab:llama2_results}
\end{table}

\section{Conclusion}
\label{sec:conclusions}

In this paper, we present FastFiD, a model based on the FiD framework, designed to accelerate the inference process for ODQA tasks. FastFiD utilizes a two-stage training technique to enable the selection of valuable sentences and focus its predictions exclusively on these sentences. Experimental results demonstrate that FastFiD substantially improves inference speed while maintaining its original answer generation performance. And our ablation study confirms the effectiveness of the two-stage training approach, showing a decrease in final performance when any single training stage is omitted.

\section*{Limitations}

The limitations of our FastFiD approach can be primarily summarized into the following two points:
\begin{itemize}
    \item Firstly, the effectiveness of our method depends on the presence of correct answers in the retrieved passages, as our approach utilizes this information to identify supported sentences. This reliance may limit its direct applicability to more complex queries. To address this issue, several strategies can be explored. For example, we can leverage the cross-attention map from FiD to identify the most informative sentences for two-stage training.
    \item Secondly, while we focus solely on the ODQA task in this paper, many other knowledge-intensive tasks also require the retrieval of numerous passages and face inference efficiency challenges. Conducting further experiments on a broader range of tasks and general RAG system will be an important avenue for future research.
\end{itemize}

\section*{Acknowledgements}
This work is supported by the National Natural Science Foundation of China (No. 62236011) and Institute Guo Qiang at Tsinghua University.

\bibliography{custom}

\begin{thebibliography}{47}
\expandafter\ifx\csname natexlab\endcsname\relax\def\natexlab#1{#1}\fi

\bibitem[{Brown et~al.(2020)Brown, Mann, Ryder, Subbiah, Kaplan, Dhariwal, Neelakantan, Shyam, Sastry, Askell, Agarwal, Herbert-Voss, Krueger, Henighan, Child, Ramesh, Ziegler, Wu, Winter, Hesse, Chen, Sigler, Litwin, Gray, Chess, Clark, Berner, McCandlish, Radford, Sutskever, and Amodei}]{gpt3-neurips}
Tom Brown, Benjamin Mann, Nick Ryder, Melanie Subbiah, Jared~D Kaplan, Prafulla Dhariwal, Arvind Neelakantan, Pranav Shyam, Girish Sastry, Amanda Askell, Sandhini Agarwal, Ariel Herbert-Voss, Gretchen Krueger, Tom Henighan, Rewon Child, Aditya Ramesh, Daniel Ziegler, Jeffrey Wu, Clemens Winter, Chris Hesse, Mark Chen, Eric Sigler, Mateusz Litwin, Scott Gray, Benjamin Chess, Jack Clark, Christopher Berner, Sam McCandlish, Alec Radford, Ilya Sutskever, and Dario Amodei. 2020.
\newblock \href {https://proceedings.neurips.cc/paper_files/paper/2020/file/1457c0d6bfcb4967418bfb8ac142f64a-Paper.pdf} {Language models are few-shot learners}.
\newblock In \emph{Advances in Neural Information Processing Systems}, volume~33, pages 1877--1901. Curran Associates, Inc.

\bibitem[{Chen et~al.(2017)Chen, Fisch, Weston, and Bordes}]{chen-etal-2017-reading}
Danqi Chen, Adam Fisch, Jason Weston, and Antoine Bordes. 2017.
\newblock \href {https://doi.org/10.18653/v1/P17-1171} {Reading {W}ikipedia to answer open-domain questions}.
\newblock In \emph{Proceedings of the 55th Annual Meeting of the Association for Computational Linguistics (Volume 1: Long Papers)}, pages 1870--1879, Vancouver, Canada. Association for Computational Linguistics.

\bibitem[{Cheng et~al.(2020)Cheng, Chang, Lee, and Toutanova}]{cheng-etal-2020-probabilistic}
Hao Cheng, Ming-Wei Chang, Kenton Lee, and Kristina Toutanova. 2020.
\newblock \href {https://doi.org/10.18653/v1/2020.acl-main.501} {Probabilistic assumptions matter: Improved models for distantly-supervised document-level question answering}.
\newblock In \emph{Proceedings of the 58th Annual Meeting of the Association for Computational Linguistics}, pages 5657--5667, Online. Association for Computational Linguistics.

\bibitem[{Cheng et~al.(2021)Cheng, Shen, Liu, He, Chen, and Gao}]{cheng-etal-2021-unitedqa}
Hao Cheng, Yelong Shen, Xiaodong Liu, Pengcheng He, Weizhu Chen, and Jianfeng Gao. 2021.
\newblock \href {https://doi.org/10.18653/v1/2021.acl-long.240} {{U}nited{QA}: {A} hybrid approach for open domain question answering}.
\newblock In \emph{Proceedings of the 59th Annual Meeting of the Association for Computational Linguistics and the 11th International Joint Conference on Natural Language Processing (Volume 1: Long Papers)}, pages 3080--3090, Online. Association for Computational Linguistics.

\bibitem[{de~Jong et~al.(2023)de~Jong, Zemlyanskiy, Ainslie, FitzGerald, Sanghai, Sha, and Cohen}]{de-jong-etal-2023-fido}
Michiel de~Jong, Yury Zemlyanskiy, Joshua Ainslie, Nicholas FitzGerald, Sumit Sanghai, Fei Sha, and William Cohen. 2023.
\newblock \href {https://doi.org/10.18653/v1/2023.findings-acl.732} {{F}i{DO}: Fusion-in-decoder optimized for stronger performance and faster inference}.
\newblock In \emph{Findings of the Association for Computational Linguistics: ACL 2023}, pages 11534--11547, Toronto, Canada. Association for Computational Linguistics.

\bibitem[{Devlin et~al.(2019)Devlin, Chang, Lee, and Toutanova}]{devlin-etal-2019-bert}
Jacob Devlin, Ming-Wei Chang, Kenton Lee, and Kristina Toutanova. 2019.
\newblock \href {https://doi.org/10.18653/v1/N19-1423} {{BERT}: Pre-training of deep bidirectional transformers for language understanding}.
\newblock In \emph{Proceedings of the 2019 Conference of the North {A}merican Chapter of the Association for Computational Linguistics: Human Language Technologies, Volume 1 (Long and Short Papers)}, pages 4171--4186, Minneapolis, Minnesota. Association for Computational Linguistics.

\bibitem[{Fajcik et~al.(2021)Fajcik, Docekal, Ondrej, and Smrz}]{fajcik-etal-2021-r2-d2}
Martin Fajcik, Martin Docekal, Karel Ondrej, and Pavel Smrz. 2021.
\newblock \href {https://doi.org/10.18653/v1/2021.findings-emnlp.73} {{R2-D2}: A modular baseline for open-domain question answering}.
\newblock In \emph{Findings of the Association for Computational Linguistics: EMNLP 2021}, pages 854--870, Punta Cana, Dominican Republic. Association for Computational Linguistics.

\bibitem[{Garg et~al.(2020)Garg, Vu, and Moschitti}]{Garg_Vu_Moschitti_2020}
Siddhant Garg, Thuy Vu, and Alessandro Moschitti. 2020.
\newblock \href {https://doi.org/10.1609/aaai.v34i05.6282} {Tanda: Transfer and adapt pre-trained transformer models for answer sentence selection}.
\newblock \emph{Proceedings of the AAAI Conference on Artificial Intelligence}, 34(05):7780--7788.

\bibitem[{Guu et~al.(2020)Guu, Lee, Tung, Pasupat, and Chang}]{pmlr-v119-guu20a}
Kelvin Guu, Kenton Lee, Zora Tung, Panupong Pasupat, and Mingwei Chang. 2020.
\newblock \href {https://proceedings.mlr.press/v119/guu20a.html} {Retrieval augmented language model pre-training}.
\newblock In \emph{Proceedings of the 37th International Conference on Machine Learning}, volume 119 of \emph{Proceedings of Machine Learning Research}, pages 3929--3938. PMLR.

\bibitem[{Hofst{\"a}tter et~al.(2023)Hofst{\"a}tter, Chen, Raman, and Zamani}]{hofstatter2023fid}
Sebastian Hofst{\"a}tter, Jiecao Chen, Karthik Raman, and Hamed Zamani. 2023.
\newblock Fid-light: Efficient and effective retrieval-augmented text generation.
\newblock In \emph{Proceedings of the 46th International ACM SIGIR Conference on Research and Development in Information Retrieval}, pages 1437--1447.

\bibitem[{Hsu et~al.(2021)Hsu, Lind, Soldaini, and Moschitti}]{hsu-etal-2021-answer}
Chao-Chun Hsu, Eric Lind, Luca Soldaini, and Alessandro Moschitti. 2021.
\newblock \href {https://doi.org/10.18653/v1/2021.findings-acl.374} {Answer generation for retrieval-based question answering systems}.
\newblock In \emph{Findings of the Association for Computational Linguistics: ACL-IJCNLP 2021}, pages 4276--4282, Online. Association for Computational Linguistics.

\bibitem[{Izacard and Grave(2021{\natexlab{a}})}]{izacard2021distilling}
Gautier Izacard and Edouard Grave. 2021{\natexlab{a}}.
\newblock \href {https://openreview.net/forum?id=NTEz-6wysdb} {Distilling knowledge from reader to retriever for question answering}.
\newblock In \emph{International Conference on Learning Representations}.

\bibitem[{Izacard and Grave(2021{\natexlab{b}})}]{izacard-grave-2021-leveraging}
Gautier Izacard and Edouard Grave. 2021{\natexlab{b}}.
\newblock \href {https://doi.org/10.18653/v1/2021.eacl-main.74} {Leveraging passage retrieval with generative models for open domain question answering}.
\newblock In \emph{Proceedings of the 16th Conference of the European Chapter of the Association for Computational Linguistics: Main Volume}, pages 874--880, Online. Association for Computational Linguistics.

\bibitem[{Jiang et~al.(2022)Jiang, Gao, Wang, Araki, Ding, Callan, and Neubig}]{jiang-etal-2022-retrieval}
Zhengbao Jiang, Luyu Gao, Zhiruo Wang, Jun Araki, Haibo Ding, Jamie Callan, and Graham Neubig. 2022.
\newblock \href {https://aclanthology.org/2022.emnlp-main.149} {Retrieval as attention: End-to-end learning of retrieval and reading within a single transformer}.
\newblock In \emph{Proceedings of the 2022 Conference on Empirical Methods in Natural Language Processing}, pages 2336--2349, Abu Dhabi, United Arab Emirates. Association for Computational Linguistics.

\bibitem[{Johnson et~al.(2021)Johnson, Douze, and Jégou}]{johnson-ann-2021}
Jeff Johnson, Matthijs Douze, and Hervé Jégou. 2021.
\newblock \href {https://doi.org/10.1109/TBDATA.2019.2921572} {Billion-scale similarity search with gpus}.
\newblock \emph{IEEE Transactions on Big Data}, 7(3):535--547.

\bibitem[{Joshi et~al.(2017)Joshi, Choi, Weld, and Zettlemoyer}]{joshi-etal-2017-triviaqa}
Mandar Joshi, Eunsol Choi, Daniel Weld, and Luke Zettlemoyer. 2017.
\newblock \href {https://doi.org/10.18653/v1/P17-1147} {{T}rivia{QA}: A large scale distantly supervised challenge dataset for reading comprehension}.
\newblock In \emph{Proceedings of the 55th Annual Meeting of the Association for Computational Linguistics (Volume 1: Long Papers)}, pages 1601--1611, Vancouver, Canada. Association for Computational Linguistics.

\bibitem[{Karpukhin et~al.(2020)Karpukhin, Oguz, Min, Lewis, Wu, Edunov, Chen, and Yih}]{karpukhin-etal-2020-dense}
Vladimir Karpukhin, Barlas Oguz, Sewon Min, Patrick Lewis, Ledell Wu, Sergey Edunov, Danqi Chen, and Wen-tau Yih. 2020.
\newblock \href {https://doi.org/10.18653/v1/2020.emnlp-main.550} {Dense passage retrieval for open-domain question answering}.
\newblock In \emph{Proceedings of the 2020 Conference on Empirical Methods in Natural Language Processing (EMNLP)}, pages 6769--6781, Online. Association for Computational Linguistics.

\bibitem[{Kwiatkowski et~al.(2019)Kwiatkowski, Palomaki, Redfield, Collins, Parikh, Alberti, Epstein, Polosukhin, Devlin, Lee, Toutanova, Jones, Kelcey, Chang, Dai, Uszkoreit, Le, and Petrov}]{kwiatkowski-etal-2019-natural}
Tom Kwiatkowski, Jennimaria Palomaki, Olivia Redfield, Michael Collins, Ankur Parikh, Chris Alberti, Danielle Epstein, Illia Polosukhin, Jacob Devlin, Kenton Lee, Kristina Toutanova, Llion Jones, Matthew Kelcey, Ming-Wei Chang, Andrew~M. Dai, Jakob Uszkoreit, Quoc Le, and Slav Petrov. 2019.
\newblock \href {https://doi.org/10.1162/tacl_a_00276} {Natural questions: A benchmark for question answering research}.
\newblock \emph{Transactions of the Association for Computational Linguistics}, 7:452--466.

\bibitem[{Lee et~al.(2022)Lee, Kedia, Lee, Paranjape, Manning, and Woo}]{lee-etal-2022-need}
Haejun Lee, Akhil Kedia, Jongwon Lee, Ashwin Paranjape, Christopher Manning, and Kyoung-Gu Woo. 2022.
\newblock \href {https://aclanthology.org/2022.emnlp-main.198} {You only need one model for open-domain question answering}.
\newblock In \emph{Proceedings of the 2022 Conference on Empirical Methods in Natural Language Processing}, pages 3047--3060, Abu Dhabi, United Arab Emirates. Association for Computational Linguistics.

\bibitem[{Lee et~al.(2021)Lee, Sung, Kang, and Chen}]{lee-etal-2021-learning-dense}
Jinhyuk Lee, Mujeen Sung, Jaewoo Kang, and Danqi Chen. 2021.
\newblock \href {https://doi.org/10.18653/v1/2021.acl-long.518} {Learning dense representations of phrases at scale}.
\newblock In \emph{Proceedings of the 59th Annual Meeting of the Association for Computational Linguistics and the 11th International Joint Conference on Natural Language Processing (Volume 1: Long Papers)}, pages 6634--6647, Online. Association for Computational Linguistics.

\bibitem[{Lee et~al.(2019)Lee, Chang, and Toutanova}]{lee-etal-2019-latent}
Kenton Lee, Ming-Wei Chang, and Kristina Toutanova. 2019.
\newblock \href {https://doi.org/10.18653/v1/P19-1612} {Latent retrieval for weakly supervised open domain question answering}.
\newblock In \emph{Proceedings of the 57th Annual Meeting of the Association for Computational Linguistics}, pages 6086--6096, Florence, Italy. Association for Computational Linguistics.

\bibitem[{Lewis et~al.(2020)Lewis, Perez, Piktus, Petroni, Karpukhin, Goyal, K\"{u}ttler, Lewis, Yih, Rockt\"{a}schel, Riedel, and Kiela}]{rag-neurips-2020}
Patrick Lewis, Ethan Perez, Aleksandra Piktus, Fabio Petroni, Vladimir Karpukhin, Naman Goyal, Heinrich K\"{u}ttler, Mike Lewis, Wen-tau Yih, Tim Rockt\"{a}schel, Sebastian Riedel, and Douwe Kiela. 2020.
\newblock \href {https://proceedings.neurips.cc/paper_files/paper/2020/file/6b493230205f780e1bc26945df7481e5-Paper.pdf} {Retrieval-augmented generation for knowledge-intensive nlp tasks}.
\newblock In \emph{Advances in Neural Information Processing Systems}, volume~33, pages 9459--9474. Curran Associates, Inc.

\bibitem[{Lewis et~al.(2021)Lewis, Wu, Liu, Minervini, Küttler, Piktus, Stenetorp, and Riedel}]{lewis-paq}
Patrick Lewis, Yuxiang Wu, Linqing Liu, Pasquale Minervini, Heinrich Küttler, Aleksandra Piktus, Pontus Stenetorp, and Sebastian Riedel. 2021.
\newblock \href {https://doi.org/10.1162/tacl_a_00415} {{PAQ: 65 Million Probably-Asked Questions and What You Can Do With Them}}.
\newblock \emph{Transactions of the Association for Computational Linguistics}, 9:1098--1115.

\bibitem[{Loshchilov and Hutter(2019)}]{DBLP:conf/iclr/LoshchilovH19}
Ilya Loshchilov and Frank Hutter. 2019.
\newblock \href {https://openreview.net/forum?id=Bkg6RiCqY7} {Decoupled weight decay regularization}.
\newblock In \emph{7th International Conference on Learning Representations, {ICLR} 2019, New Orleans, LA, USA, May 6-9, 2019}. OpenReview.net.

\bibitem[{Malkov and Yashunin(2020)}]{malkov-hsnw-2020}
Yu~A. Malkov and D.~A. Yashunin. 2020.
\newblock \href {https://doi.org/10.1109/TPAMI.2018.2889473} {Efficient and robust approximate nearest neighbor search using hierarchical navigable small world graphs}.
\newblock \emph{IEEE Transactions on Pattern Analysis and Machine Intelligence}, 42(4):824--836.

\bibitem[{Matsubara et~al.(2020)Matsubara, Vu, and Moschitti}]{10.1145/3397271.3401266}
Yoshitomo Matsubara, Thuy Vu, and Alessandro Moschitti. 2020.
\newblock \href {https://doi.org/10.1145/3397271.3401266} {Reranking for efficient transformer-based answer selection}.
\newblock In \emph{Proceedings of the 43rd International ACM SIGIR Conference on Research and Development in Information Retrieval}, SIGIR '20, page 1577–1580, New York, NY, USA. Association for Computing Machinery.

\bibitem[{Min et~al.(2019)Min, Chen, Hajishirzi, and Zettlemoyer}]{min-etal-2019-discrete}
Sewon Min, Danqi Chen, Hannaneh Hajishirzi, and Luke Zettlemoyer. 2019.
\newblock \href {https://doi.org/10.18653/v1/D19-1284} {A discrete hard {EM} approach for weakly supervised question answering}.
\newblock In \emph{Proceedings of the 2019 Conference on Empirical Methods in Natural Language Processing and the 9th International Joint Conference on Natural Language Processing (EMNLP-IJCNLP)}, pages 2851--2864, Hong Kong, China. Association for Computational Linguistics.

\bibitem[{Min et~al.(2020)Min, Michael, Hajishirzi, and Zettlemoyer}]{min-etal-2020-ambigqa}
Sewon Min, Julian Michael, Hannaneh Hajishirzi, and Luke Zettlemoyer. 2020.
\newblock \href {https://doi.org/10.18653/v1/2020.emnlp-main.466} {{A}mbig{QA}: Answering ambiguous open-domain questions}.
\newblock In \emph{Proceedings of the 2020 Conference on Empirical Methods in Natural Language Processing (EMNLP)}, pages 5783--5797, Online. Association for Computational Linguistics.

\bibitem[{Nogueira and Cho(2020)}]{nogueira2020passage}
Rodrigo Nogueira and Kyunghyun Cho. 2020.
\newblock \href {http://arxiv.org/abs/1901.04085} {Passage re-ranking with bert}.

\bibitem[{OpenAI et~al.(2024)OpenAI, Achiam, Adler, Agarwal, Ahmad, Akkaya, Aleman, Almeida, Altenschmidt, Altman, Anadkat, Avila, Babuschkin, Balaji, Balcom, Baltescu, Bao, Bavarian, Belgum, Bello, Berdine, Bernadett-Shapiro, Berner, Bogdonoff, Boiko, Boyd, Brakman, Brockman, Brooks, Brundage, Button, Cai, Campbell, Cann, Carey, Carlson, Carmichael, Chan, Chang, Chantzis, Chen, Chen, Chen, Chen, Chen, Chess, Cho, Chu, Chung, Cummings, Currier, Dai, Decareaux, Degry, Deutsch, Deville, Dhar, Dohan, Dowling, Dunning, Ecoffet, Eleti, Eloundou, Farhi, Fedus, Felix, Fishman, Forte, Fulford, Gao, Georges, Gibson, Goel, Gogineni, Goh, Gontijo-Lopes, Gordon, Grafstein, Gray, Greene, Gross, Gu, Guo, Hallacy, Han, Harris, He, Heaton, Heidecke, Hesse, Hickey, Hickey, Hoeschele, Houghton, Hsu, Hu, Hu, Huizinga, Jain, Jain, Jang, Jiang, Jiang, Jin, Jin, Jomoto, Jonn, Jun, Kaftan, Łukasz Kaiser, Kamali, Kanitscheider, Keskar, Khan, Kilpatrick, Kim, Kim, Kim, Kirchner, Kiros, Knight, Kokotajlo, Łukasz Kondraciuk,
  Kondrich, Konstantinidis, Kosic, Krueger, Kuo, Lampe, Lan, Lee, Leike, Leung, Levy, Li, Lim, Lin, Lin, Litwin, Lopez, Lowe, Lue, Makanju, Malfacini, Manning, Markov, Markovski, Martin, Mayer, Mayne, McGrew, McKinney, McLeavey, McMillan, McNeil, Medina, Mehta, Menick, Metz, Mishchenko, Mishkin, Monaco, Morikawa, Mossing, Mu, Murati, Murk, Mély, Nair, Nakano, Nayak, Neelakantan, Ngo, Noh, Ouyang, O'Keefe, Pachocki, Paino, Palermo, Pantuliano, Parascandolo, Parish, Parparita, Passos, Pavlov, Peng, Perelman, de~Avila Belbute~Peres, Petrov, de~Oliveira~Pinto, Michael, Pokorny, Pokrass, Pong, Powell, Power, Power, Proehl, Puri, Radford, Rae, Ramesh, Raymond, Real, Rimbach, Ross, Rotsted, Roussez, Ryder, Saltarelli, Sanders, Santurkar, Sastry, Schmidt, Schnurr, Schulman, Selsam, Sheppard, Sherbakov, Shieh, Shoker, Shyam, Sidor, Sigler, Simens, Sitkin, Slama, Sohl, Sokolowsky, Song, Staudacher, Such, Summers, Sutskever, Tang, Tezak, Thompson, Tillet, Tootoonchian, Tseng, Tuggle, Turley, Tworek, Uribe, Vallone,
  Vijayvergiya, Voss, Wainwright, Wang, Wang, Wang, Ward, Wei, Weinmann, Welihinda, Welinder, Weng, Weng, Wiethoff, Willner, Winter, Wolrich, Wong, Workman, Wu, Wu, Wu, Xiao, Xu, Yoo, Yu, Yuan, Zaremba, Zellers, Zhang, Zhang, Zhao, Zheng, Zhuang, Zhuk, and Zoph}]{openai2024gpt4}
OpenAI, Josh Achiam, Steven Adler, Sandhini Agarwal, Lama Ahmad, Ilge Akkaya, Florencia~Leoni Aleman, Diogo Almeida, Janko Altenschmidt, Sam Altman, Shyamal Anadkat, Red Avila, Igor Babuschkin, Suchir Balaji, Valerie Balcom, Paul Baltescu, Haiming Bao, Mohammad Bavarian, Jeff Belgum, Irwan Bello, Jake Berdine, Gabriel Bernadett-Shapiro, Christopher Berner, Lenny Bogdonoff, Oleg Boiko, Madelaine Boyd, Anna-Luisa Brakman, Greg Brockman, Tim Brooks, Miles Brundage, Kevin Button, Trevor Cai, Rosie Campbell, Andrew Cann, Brittany Carey, Chelsea Carlson, Rory Carmichael, Brooke Chan, Che Chang, Fotis Chantzis, Derek Chen, Sully Chen, Ruby Chen, Jason Chen, Mark Chen, Ben Chess, Chester Cho, Casey Chu, Hyung~Won Chung, Dave Cummings, Jeremiah Currier, Yunxing Dai, Cory Decareaux, Thomas Degry, Noah Deutsch, Damien Deville, Arka Dhar, David Dohan, Steve Dowling, Sheila Dunning, Adrien Ecoffet, Atty Eleti, Tyna Eloundou, David Farhi, Liam Fedus, Niko Felix, Simón~Posada Fishman, Juston Forte, Isabella Fulford, Leo
  Gao, Elie Georges, Christian Gibson, Vik Goel, Tarun Gogineni, Gabriel Goh, Rapha Gontijo-Lopes, Jonathan Gordon, Morgan Grafstein, Scott Gray, Ryan Greene, Joshua Gross, Shixiang~Shane Gu, Yufei Guo, Chris Hallacy, Jesse Han, Jeff Harris, Yuchen He, Mike Heaton, Johannes Heidecke, Chris Hesse, Alan Hickey, Wade Hickey, Peter Hoeschele, Brandon Houghton, Kenny Hsu, Shengli Hu, Xin Hu, Joost Huizinga, Shantanu Jain, Shawn Jain, Joanne Jang, Angela Jiang, Roger Jiang, Haozhun Jin, Denny Jin, Shino Jomoto, Billie Jonn, Heewoo Jun, Tomer Kaftan, Łukasz Kaiser, Ali Kamali, Ingmar Kanitscheider, Nitish~Shirish Keskar, Tabarak Khan, Logan Kilpatrick, Jong~Wook Kim, Christina Kim, Yongjik Kim, Jan~Hendrik Kirchner, Jamie Kiros, Matt Knight, Daniel Kokotajlo, Łukasz Kondraciuk, Andrew Kondrich, Aris Konstantinidis, Kyle Kosic, Gretchen Krueger, Vishal Kuo, Michael Lampe, Ikai Lan, Teddy Lee, Jan Leike, Jade Leung, Daniel Levy, Chak~Ming Li, Rachel Lim, Molly Lin, Stephanie Lin, Mateusz Litwin, Theresa Lopez, Ryan
  Lowe, Patricia Lue, Anna Makanju, Kim Malfacini, Sam Manning, Todor Markov, Yaniv Markovski, Bianca Martin, Katie Mayer, Andrew Mayne, Bob McGrew, Scott~Mayer McKinney, Christine McLeavey, Paul McMillan, Jake McNeil, David Medina, Aalok Mehta, Jacob Menick, Luke Metz, Andrey Mishchenko, Pamela Mishkin, Vinnie Monaco, Evan Morikawa, Daniel Mossing, Tong Mu, Mira Murati, Oleg Murk, David Mély, Ashvin Nair, Reiichiro Nakano, Rajeev Nayak, Arvind Neelakantan, Richard Ngo, Hyeonwoo Noh, Long Ouyang, Cullen O'Keefe, Jakub Pachocki, Alex Paino, Joe Palermo, Ashley Pantuliano, Giambattista Parascandolo, Joel Parish, Emy Parparita, Alex Passos, Mikhail Pavlov, Andrew Peng, Adam Perelman, Filipe de~Avila Belbute~Peres, Michael Petrov, Henrique~Ponde de~Oliveira~Pinto, Michael, Pokorny, Michelle Pokrass, Vitchyr~H. Pong, Tolly Powell, Alethea Power, Boris Power, Elizabeth Proehl, Raul Puri, Alec Radford, Jack Rae, Aditya Ramesh, Cameron Raymond, Francis Real, Kendra Rimbach, Carl Ross, Bob Rotsted, Henri Roussez,
  Nick Ryder, Mario Saltarelli, Ted Sanders, Shibani Santurkar, Girish Sastry, Heather Schmidt, David Schnurr, John Schulman, Daniel Selsam, Kyla Sheppard, Toki Sherbakov, Jessica Shieh, Sarah Shoker, Pranav Shyam, Szymon Sidor, Eric Sigler, Maddie Simens, Jordan Sitkin, Katarina Slama, Ian Sohl, Benjamin Sokolowsky, Yang Song, Natalie Staudacher, Felipe~Petroski Such, Natalie Summers, Ilya Sutskever, Jie Tang, Nikolas Tezak, Madeleine~B. Thompson, Phil Tillet, Amin Tootoonchian, Elizabeth Tseng, Preston Tuggle, Nick Turley, Jerry Tworek, Juan Felipe~Cerón Uribe, Andrea Vallone, Arun Vijayvergiya, Chelsea Voss, Carroll Wainwright, Justin~Jay Wang, Alvin Wang, Ben Wang, Jonathan Ward, Jason Wei, CJ~Weinmann, Akila Welihinda, Peter Welinder, Jiayi Weng, Lilian Weng, Matt Wiethoff, Dave Willner, Clemens Winter, Samuel Wolrich, Hannah Wong, Lauren Workman, Sherwin Wu, Jeff Wu, Michael Wu, Kai Xiao, Tao Xu, Sarah Yoo, Kevin Yu, Qiming Yuan, Wojciech Zaremba, Rowan Zellers, Chong Zhang, Marvin Zhang, Shengjia
  Zhao, Tianhao Zheng, Juntang Zhuang, William Zhuk, and Barret Zoph. 2024.
\newblock \href {http://arxiv.org/abs/2303.08774} {Gpt-4 technical report}.

\bibitem[{Paszke et~al.(2019)Paszke, Gross, Massa, Lerer, Bradbury, Chanan, Killeen, Lin, Gimelshein, Antiga, Desmaison, Kopf, Yang, DeVito, Raison, Tejani, Chilamkurthy, Steiner, Fang, Bai, and Chintala}]{adam-etal-2019-pytorch}
Adam Paszke, Sam Gross, Francisco Massa, Adam Lerer, James Bradbury, Gregory Chanan, Trevor Killeen, Zeming Lin, Natalia Gimelshein, Luca Antiga, Alban Desmaison, Andreas Kopf, Edward Yang, Zachary DeVito, Martin Raison, Alykhan Tejani, Sasank Chilamkurthy, Benoit Steiner, Lu~Fang, Junjie Bai, and Soumith Chintala. 2019.
\newblock \href {https://proceedings.neurips.cc/paper_files/paper/2019/file/bdbca288fee7f92f2bfa9f7012727740-Paper.pdf} {Pytorch: An imperative style, high-performance deep learning library}.
\newblock In \emph{Advances in Neural Information Processing Systems}, volume~32. Curran Associates, Inc.

\bibitem[{Raffel et~al.(2020)Raffel, Shazeer, Roberts, Lee, Narang, Matena, Zhou, Li, and Liu}]{t5-jmlr}
Colin Raffel, Noam Shazeer, Adam Roberts, Katherine Lee, Sharan Narang, Michael Matena, Yanqi Zhou, Wei Li, and Peter~J. Liu. 2020.
\newblock \href {http://jmlr.org/papers/v21/20-074.html} {Exploring the limits of transfer learning with a unified text-to-text transformer}.
\newblock \emph{Journal of Machine Learning Research}, 21(140):1--67.

\bibitem[{Rajbhandari et~al.(2020)Rajbhandari, Rasley, Ruwase, and He}]{rajbhandari2020zero}
Samyam Rajbhandari, Jeff Rasley, Olatunji Ruwase, and Yuxiong He. 2020.
\newblock Zero: Memory optimizations toward training trillion parameter models.
\newblock ArXiv.

\bibitem[{Roberts et~al.(2020)Roberts, Raffel, and Shazeer}]{roberts-etal-2020-much}
Adam Roberts, Colin Raffel, and Noam Shazeer. 2020.
\newblock \href {https://doi.org/10.18653/v1/2020.emnlp-main.437} {How much knowledge can you pack into the parameters of a language model?}
\newblock In \emph{Proceedings of the 2020 Conference on Empirical Methods in Natural Language Processing (EMNLP)}, pages 5418--5426, Online. Association for Computational Linguistics.

\bibitem[{Sachan et~al.(2021{\natexlab{a}})Sachan, Patwary, Shoeybi, Kant, Ping, Hamilton, and Catanzaro}]{sachan-etal-2021-end}
Devendra Sachan, Mostofa Patwary, Mohammad Shoeybi, Neel Kant, Wei Ping, William~L. Hamilton, and Bryan Catanzaro. 2021{\natexlab{a}}.
\newblock \href {https://doi.org/10.18653/v1/2021.acl-long.519} {End-to-end training of neural retrievers for open-domain question answering}.
\newblock In \emph{Proceedings of the 59th Annual Meeting of the Association for Computational Linguistics and the 11th International Joint Conference on Natural Language Processing (Volume 1: Long Papers)}, pages 6648--6662, Online. Association for Computational Linguistics.

\bibitem[{Sachan et~al.(2021{\natexlab{b}})Sachan, Reddy, Hamilton, Dyer, and Yogatama}]{sachan2021endtoend}
Devendra~Singh Sachan, Siva Reddy, William~L. Hamilton, Chris Dyer, and Dani Yogatama. 2021{\natexlab{b}}.
\newblock \href {https://openreview.net/forum?id=5KWmB6JePx} {End-to-end training of multi-document reader and retriever for open-domain question answering}.
\newblock In \emph{Advances in Neural Information Processing Systems}.

\bibitem[{Seo et~al.(2019)Seo, Lee, Kwiatkowski, Parikh, Farhadi, and Hajishirzi}]{seo-etal-2019-real}
Minjoon Seo, Jinhyuk Lee, Tom Kwiatkowski, Ankur Parikh, Ali Farhadi, and Hannaneh Hajishirzi. 2019.
\newblock \href {https://doi.org/10.18653/v1/P19-1436} {Real-time open-domain question answering with dense-sparse phrase index}.
\newblock In \emph{Proceedings of the 57th Annual Meeting of the Association for Computational Linguistics}, pages 4430--4441, Florence, Italy. Association for Computational Linguistics.

\bibitem[{Severyn and Moschitti(2015)}]{10.1145/2766462.2767738}
Aliaksei Severyn and Alessandro Moschitti. 2015.
\newblock \href {https://doi.org/10.1145/2766462.2767738} {Learning to rank short text pairs with convolutional deep neural networks}.
\newblock In \emph{Proceedings of the 38th International ACM SIGIR Conference on Research and Development in Information Retrieval}, SIGIR '15, page 373–382, New York, NY, USA. Association for Computing Machinery.

\bibitem[{Shen et~al.(2017)Shen, Yang, and Deng}]{shen-etal-2017-inter}
Gehui Shen, Yunlun Yang, and Zhi-Hong Deng. 2017.
\newblock \href {https://doi.org/10.18653/v1/D17-1122} {Inter-weighted alignment network for sentence pair modeling}.
\newblock In \emph{Proceedings of the 2017 Conference on Empirical Methods in Natural Language Processing}, pages 1179--1189, Copenhagen, Denmark. Association for Computational Linguistics.

\bibitem[{Stelmakh et~al.(2023)Stelmakh, Luan, Dhingra, and Chang}]{stelmakh2023asqa}
Ivan Stelmakh, Yi~Luan, Bhuwan Dhingra, and Ming-Wei Chang. 2023.
\newblock \href {http://arxiv.org/abs/2204.06092} {Asqa: Factoid questions meet long-form answers}.

\bibitem[{Touvron et~al.(2023{\natexlab{a}})Touvron, Lavril, Izacard, Martinet, Lachaux, Lacroix, Rozi{\`{e}}re, Goyal, Hambro, Azhar, Rodriguez, Joulin, Grave, and Lample}]{DBLP:journals/corr/abs-2302-13971}
Hugo Touvron, Thibaut Lavril, Gautier Izacard, Xavier Martinet, Marie{-}Anne Lachaux, Timoth{\'{e}}e Lacroix, Baptiste Rozi{\`{e}}re, Naman Goyal, Eric Hambro, Faisal Azhar, Aur{\'{e}}lien Rodriguez, Armand Joulin, Edouard Grave, and Guillaume Lample. 2023{\natexlab{a}}.
\newblock \href {https://doi.org/10.48550/ARXIV.2302.13971} {Llama: Open and efficient foundation language models}.
\newblock \emph{CoRR}, abs/2302.13971.

\bibitem[{Touvron et~al.(2023{\natexlab{b}})Touvron, Martin, Stone, Albert, Almahairi, Babaei, Bashlykov, Batra, Bhargava, Bhosale, Bikel, Blecher, Canton{-}Ferrer, Chen, Cucurull, Esiobu, Fernandes, Fu, Fu, Fuller, Gao, Goswami, Goyal, Hartshorn, Hosseini, Hou, Inan, Kardas, Kerkez, Khabsa, Kloumann, Korenev, Koura, Lachaux, Lavril, Lee, Liskovich, Lu, Mao, Martinet, Mihaylov, Mishra, Molybog, Nie, Poulton, Reizenstein, Rungta, Saladi, Schelten, Silva, Smith, Subramanian, Tan, Tang, Taylor, Williams, Kuan, Xu, Yan, Zarov, Zhang, Fan, Kambadur, Narang, Rodriguez, Stojnic, Edunov, and Scialom}]{DBLP:journals/corr/abs-2307-09288}
Hugo Touvron, Louis Martin, Kevin Stone, Peter Albert, Amjad Almahairi, Yasmine Babaei, Nikolay Bashlykov, Soumya Batra, Prajjwal Bhargava, Shruti Bhosale, Dan Bikel, Lukas Blecher, Cristian Canton{-}Ferrer, Moya Chen, Guillem Cucurull, David Esiobu, Jude Fernandes, Jeremy Fu, Wenyin Fu, Brian Fuller, Cynthia Gao, Vedanuj Goswami, Naman Goyal, Anthony Hartshorn, Saghar Hosseini, Rui Hou, Hakan Inan, Marcin Kardas, Viktor Kerkez, Madian Khabsa, Isabel Kloumann, Artem Korenev, Punit~Singh Koura, Marie{-}Anne Lachaux, Thibaut Lavril, Jenya Lee, Diana Liskovich, Yinghai Lu, Yuning Mao, Xavier Martinet, Todor Mihaylov, Pushkar Mishra, Igor Molybog, Yixin Nie, Andrew Poulton, Jeremy Reizenstein, Rashi Rungta, Kalyan Saladi, Alan Schelten, Ruan Silva, Eric~Michael Smith, Ranjan Subramanian, Xiaoqing~Ellen Tan, Binh Tang, Ross Taylor, Adina Williams, Jian~Xiang Kuan, Puxin Xu, Zheng Yan, Iliyan Zarov, Yuchen Zhang, Angela Fan, Melanie Kambadur, Sharan Narang, Aur{\'{e}}lien Rodriguez, Robert Stojnic, Sergey Edunov,
  and Thomas Scialom. 2023{\natexlab{b}}.
\newblock \href {https://doi.org/10.48550/ARXIV.2307.09288} {Llama 2: Open foundation and fine-tuned chat models}.
\newblock \emph{CoRR}, abs/2307.09288.

\bibitem[{Wen et~al.(2022)Wen, Wang, Luo, and Wang}]{wen-etal-2022-m3}
Liang Wen, Houfeng Wang, Yingwei Luo, and Xiaolin Wang. 2022.
\newblock \href {https://aclanthology.org/2022.emnlp-main.94} {{M}3: A multi-view fusion and multi-decoding network for multi-document reading comprehension}.
\newblock In \emph{Proceedings of the 2022 Conference on Empirical Methods in Natural Language Processing}, pages 1450--1461, Abu Dhabi, United Arab Emirates. Association for Computational Linguistics.

\bibitem[{Wolf et~al.(2020)Wolf, Debut, Sanh, Chaumond, Delangue, Moi, Cistac, Rault, Louf, Funtowicz, Davison, Shleifer, von Platen, Ma, Jernite, Plu, Xu, Le~Scao, Gugger, Drame, Lhoest, and Rush}]{wolf-etal-2020-transformers}
Thomas Wolf, Lysandre Debut, Victor Sanh, Julien Chaumond, Clement Delangue, Anthony Moi, Pierric Cistac, Tim Rault, Remi Louf, Morgan Funtowicz, Joe Davison, Sam Shleifer, Patrick von Platen, Clara Ma, Yacine Jernite, Julien Plu, Canwen Xu, Teven Le~Scao, Sylvain Gugger, Mariama Drame, Quentin Lhoest, and Alexander Rush. 2020.
\newblock \href {https://doi.org/10.18653/v1/2020.emnlp-demos.6} {Transformers: State-of-the-art natural language processing}.
\newblock In \emph{Proceedings of the 2020 Conference on Empirical Methods in Natural Language Processing: System Demonstrations}, pages 38--45, Online. Association for Computational Linguistics.

\bibitem[{Wu et~al.(2021)Wu, Minervini, Stenetorp, and Riedel}]{wu-etal-2021-training}
Yuxiang Wu, Pasquale Minervini, Pontus Stenetorp, and Sebastian Riedel. 2021.
\newblock \href {https://doi.org/10.18653/v1/2021.acl-short.57} {Training adaptive computation for open-domain question answering with computational constraints}.
\newblock In \emph{Proceedings of the 59th Annual Meeting of the Association for Computational Linguistics and the 11th International Joint Conference on Natural Language Processing (Volume 2: Short Papers)}, pages 447--453, Online. Association for Computational Linguistics.

\bibitem[{Wu et~al.(2020)Wu, Riedel, Minervini, and Stenetorp}]{wu-etal-2020-dont}
Yuxiang Wu, Sebastian Riedel, Pasquale Minervini, and Pontus Stenetorp. 2020.
\newblock \href {https://doi.org/10.18653/v1/2020.emnlp-main.244} {Don{'}t read too much into it: Adaptive computation for open-domain question answering}.
\newblock In \emph{Proceedings of the 2020 Conference on Empirical Methods in Natural Language Processing (EMNLP)}, pages 3029--3039, Online. Association for Computational Linguistics.

\bibitem[{Yu et~al.(2022)Yu, Zhu, Fang, Yu, Wang, Xu, Ren, Yang, and Zeng}]{yu-etal-2022-kg}
Donghan Yu, Chenguang Zhu, Yuwei Fang, Wenhao Yu, Shuohang Wang, Yichong Xu, Xiang Ren, Yiming Yang, and Michael Zeng. 2022.
\newblock \href {https://doi.org/10.18653/v1/2022.acl-long.340} {{KG}-{F}i{D}: Infusing knowledge graph in fusion-in-decoder for open-domain question answering}.
\newblock In \emph{Proceedings of the 60th Annual Meeting of the Association for Computational Linguistics (Volume 1: Long Papers)}, pages 4961--4974, Dublin, Ireland. Association for Computational Linguistics.

\end{thebibliography}

\appendix

\section*{Appendices}
\label{sec:appendix}

\section{Implementation Details}
\label{sec:implemenration}

In the first stage of training, we employ a linear scheduler with a warmup ratio of $0.1$ and a maximum learning rate of $10^{-4}$ for $10$ epochs. The selection of the best checkpoint for the second-stage training is based on performance evaluation on the development set. In the second training stage, we use a constant learning rate of $5 \times 10^{-5}$ for $5$ epochs. We evaluate the performance of the hyperparameter $\lambda$ in the training objective using values of $0.1$ and $0.05$, and select the one that yields better results for each dataset. Specifically, we use $0.1$ for NQ and ASQA, and $0.05$ for TriviaQA, considering its higher number of annotated sentences as indicated in Table~\ref{tab:statistics}.

During inference, we follow the approach of previous work~\citep{hofstatter2023fid} by utilizing beam search with a beam size of $4$. The maximum decoding length is set to $32$ for NQ and TriviaQA, while it is set to $128$ for ASQA due to the longer answer lengths in that dataset.

\section{Case Study}
\label{sec:case_study}

\begin{figure*}[t]
    \centering
    \includegraphics[width=\linewidth]{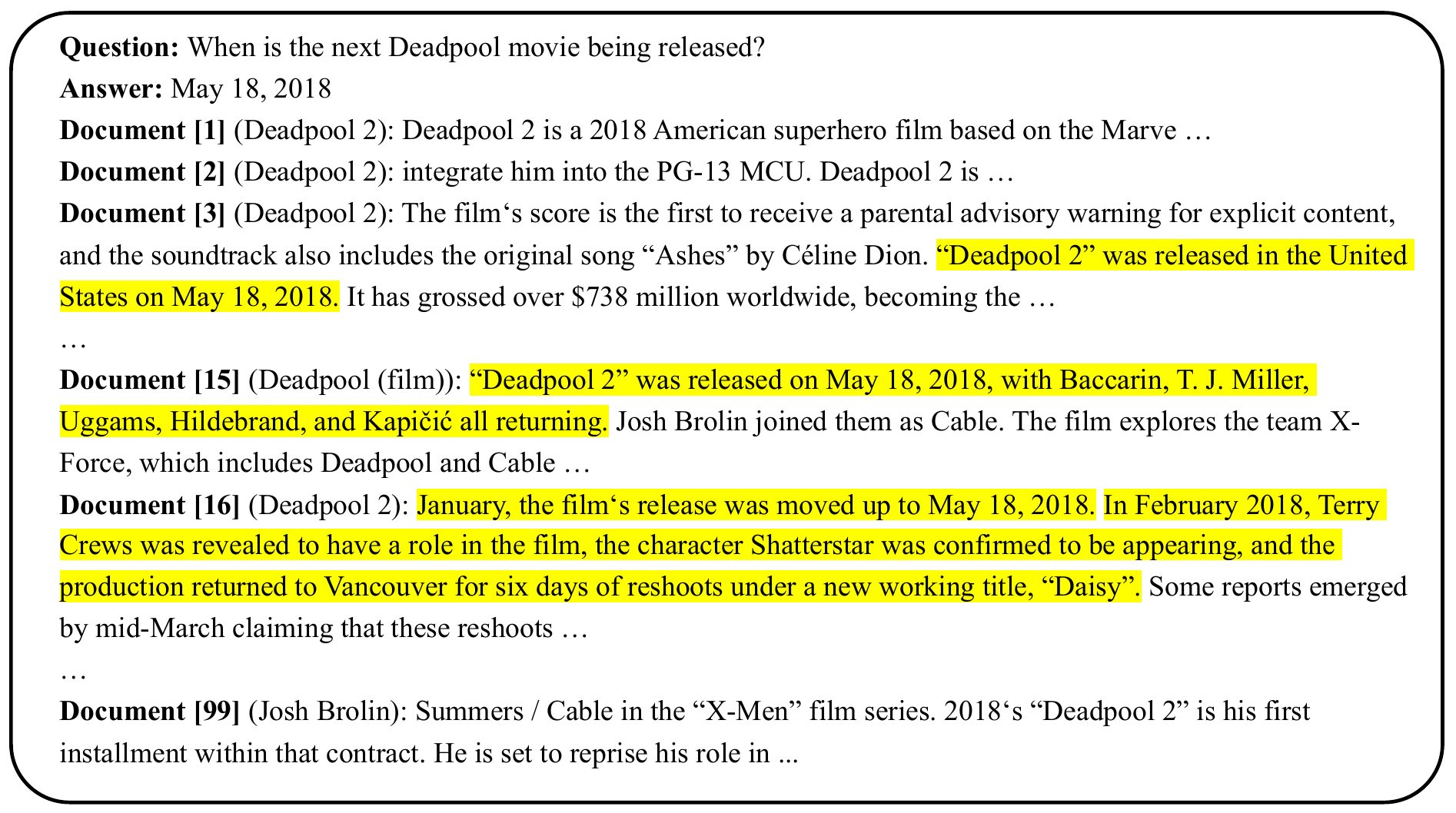}
    \caption{An example from the test set of NQ with $100$ retrieved passages. The text highlighted in yellow represents the valuable sentences identified by our FastFiD.}
    \label{fig:case}
\end{figure*}

To demonstrate the effectiveness of our FastFiD approach, we present an example using the test set of NQ, as depicted in Figure~\ref{fig:case}. In this figure, the text highlighted in yellow represents the valuable sentences identified by FastFiD, which are subsequently utilized in the decoding process. It is evident that FastFiD possesses the capability to recognize valuable sentences that often contain the correct answer, even if they are not in the highly-ranked documents. Additionally, these valuable sentences only constitute a small portion of all the retrieved passages which is important for us to accelerate inference. However, it is important to note that not all selected sentences are necessarily relevant to the given question. For instance, the second selected sentence in \textsc{Document [16]} may not carry any meaningful information. Consequently, we need to select a specific number of sentences to retain all the pertinent information for achieving satisfactory performance, as demonstrated in Section~\ref{subsec:number_sentences}.

\section{Ablation Study of Selected Sentences Number on TriviaQA}
\label{subsec:sentence_num_tqa}

\begin{table}[t]
\small
\begin{tabular}{@{}l|ccc@{}}
\toprule
\multicolumn{1}{c|}{\textbf{Model}} & \textbf{\begin{tabular}[c]{@{}c@{}}\# Select\\ Sentence\end{tabular}} & \textbf{TriviaQA-Test} & \textbf{\begin{tabular}[c]{@{}c@{}}Context\\ Length\end{tabular}} \\ \midrule
FiD      & -    & \textbf{69.79}     & 38,400  \\ \hline
FastFiD  & 50  & 67.57 & 487   \\
FastFiD   & 100  & 67.96 & 723  \\
FastFiD  & 200 &  68.71 & 1,449  \\
FastFiD  & 400 & 69.35 & 2,933  \\
FastFiD  & 800 & 69.56 & 5,038  \\ \bottomrule
\end{tabular}
\caption{Experiments on the number of selected sentences on TriviaQA.}
\label{tab:number_of_select_tqa}
\end{table}

In Section~\ref{subsec:number_sentences}, we examine the impact of varying the number of selected sentences on the final performance using the NQ dataset. To determine if the optimal number of selected sentences varies across different datasets, we extend our experiments to TriviaQA.

The results, presented in Table~\ref{tab:number_of_select_tqa}, reveal a different trend compared to the NQ dataset findings in Table~\ref{tab:number_of_select}. In the case of TriviaQA, performance continually improves as the number of selected sentences increases. However, the marginal gains decrease with the inclusion of more sentences. For instance, increasing the number of sentences from 200 to 400 leads to a performance improvement of 0.64, while an increase from 400 to 800 sentences results in a smaller gain of 0.21.

The observed trend can be attributed to the fact that TriviaQA contains a greater average number of supportive sentences per question compared to NQ. As shown in Table~\ref{tab:statistics}, NQ has an average of 14.84 supportive sentences, whereas TriviaQA has 30.58, nearly double the amount. Consequently, selecting more sentences in TriviaQA provides additional supportive information that is beneficial for answering the questions. Conversely, in the NQ dataset, increasing the number of selected sentences might introduce more noise, leading to incorrect answers. Therefore, we conclude that the optimal number of sentences to select may vary across datasets, depending on how concentrated the relevant information is within each dataset.

\section{Influence of Model Size}
\label{subsec:model_size}

\begin{table}[ht]
    \centering
    \begin{tabular}{lccc}
    \toprule
        \textbf{Model} & \textbf{EM} & \textbf{TPQ} & \textbf{Speed} \\ \hline
        FiD-Base & 50.06 & 514 & 1.0X \\
        FastFiD-Base & \textbf{50.17} & \textbf{148} & \textbf{3.5X} \\ \hline
        FiD-Large & \textbf{53.60} & 1,262 & 1.0X \\
        FastFiD-Large & 53.19 & \textbf{368} & \textbf{3.4X} \\ \bottomrule
    \end{tabular}
    \caption{Performance of vanilla FiD, and FastFiD with $100$ retrieved passages and different model sizes on NQ test set. We select $200$ sentences for FastFiD. TPQ is measured by milliseconds.}
    \label{tab:large_results}
\end{table}

To verify the effectiveness of our method across different model scales, we conducted additional experiments using T5-Large, which consists of 770 million parameters. The performance of various methods on T5-Large is detailed in Table~\ref{tab:large_results}. Our results demonstrate that our method remains effective on larger models, achieving a speedup of \textbf{3.4X}. Moreover, FastFiD-Large outperforms FiD-Base in both speed and the EM metric, indicating that our method allows for the utilization of larger models to enhance QA performance without increasing inference time.


\end{document}